# KLCBL: An Improved Police Incident Classification Model

Liu Zhuoxian[1]    Shi Tuo[2]    Hu Xiaofeng[1]

[1]People's Public Security University of China, School of Cyber Security, Beijing 100038, China

[2] Beijing Police College, Department of Public Security Management, Beijing 102202, China

**Abstract:** The police incident data serves as a vital source of public security intelligence, and its classification and mining are of significant importance for crime prevention and control. However, grassroots public security agencies often face challenges in efficiently classifying this data due to the inefficiency of manual methods and the poor performance of automated classification systems when dealing with complex and diverse incident types. This issue is particularly pronounced in the case of telecom and online fraud, which accounts for the largest volume of cases and presents severe difficulties for automated systems. To address this problem, there is an urgent need for a highly efficient and accurate intelligent model that can automate the classification of police incidents, thereby improving the utilization of police resources and operational effectiveness. In response to this challenge, this research proposes a multichannel neural network model named KLCBL, designed for the classification of police incident data. The model integrates the (Kolmogorov–Arnold Networks, KAN) algorithm, a linguistically enhanced text preprocessing approach (Language Representation Model, LERT), a (Convolutional Neural Network, CNN), and a (Bidirectional Long Short-Term Memory, BiLSTM) network. The LERT model is employed to extract various linguistic features, while the parallel CNN and BiLSTM structures are used to capture local semantic information and contextual sequential features, respectively. The KAN network further optimizes the neural network structure through strip fitting, enhancing the model's performance. The proposed model was evaluated using real police incident data from a northern city in China, achieving a classification accuracy of 91.9%. Comparative ablation experiments demonstrated that the KLCBL model outperforms baseline models, validating its effectiveness. This study confirms that the model successfully addresses the classification challenges posed by 110 police incident data, providing grassroots public security agencies with a powerful intelligent classification tool. The model enhances the informatization of police operations and improves the efficient allocation of police resources, aligning with practical operational needs. Furthermore, the model's multichannel architecture and integrated algorithms offer broad applicability to other classification tasks, with potential for future exploration in additional domains.

**Keywords:**   Police Incident Classification; KAN Network; Multichannel Neural Network; Telecom and Online Fraud.

## 1  Introduction

The police incident data refers to the information recorded by grassroots public security agencies during the process of receiving and responding to incidents. It typically contains the initial records of reports made by officers based on the complainant's account, providing key information and a general overview of the case[1]. Effectively and scientifically maximizing the use of incident data is crucial for enhancing crime prevention, investigation, and case-solving efforts by public security agencies[2]. One method for handling incident data is to classify it according to case types[7], such as telecom and online fraud[3], public order cases[4], and major property-related offenses[5], allowing tailored control and prevention measures for each case category[8]. However, in real-world police work, police incident text data is characterized by high volume, complex content, and inconsistent language descriptions[9], making it difficult for grassroots officers to efficiently process and analyze this data using traditional methods.

It has been reported that most grassroots public security agencies rely on manual classification of police incident data, primarily using simple rule-based matching. This approach is labor-intensive, consumes significant police resources, and increases the workload of officers. The greatest challenge lies in distinguishing telecom and online fraud incidents from other general fraud cases and other types of incidents. Telecom and online fraud incidents, due to their diverse crime types and complex manifestations, are often confused with other categories, complicating the classification process. In recent years, the rise of deep learning and artificial intelligence has prompted researchers to propose new methods for police incident text classification[6,8,9,34,37,40]. While some natural language processing models have shown promising results in classifying certain types of incidents, many models struggle with generalization, and classification accuracy, particularly for telecom and online fraud, remains inadequate for practical deployment.

To address these challenges, this paper presents a novel multichannel neural network model named KLCBL, which integrates the LERT text preprocessing method, CNN (Convolutional Neural Network), BiLSTM (Bidirectional Long Short-Term Memory) neural networks, and the KAN (Kolmogorov–Arnold Networks) algorithm. This model classifies police incidents into three categories: telecom fraud, non-telecom fraud, and general incidents. Experimental results, based on real-world data, demonstrated a classification accuracy of 91.8%, addressing the challenge of classifying telecom fraud, which represents the largest portion of police incident data. In January 2024, the model was successfully deployed in the anti-fraud department of a northern city in China. The model's flexibility also allows for its application to other incident classification tasks and broader classification scenarios.

## 2 Related Research

### 2.1 Text Classification

Text classification is one of the core tasks in natural language processing. There are several commonly used approaches to address this problem:

One approach involves fine-tuning pre-trained models for specific downstream tasks[10,13]. Pre-trained models, which are trained on large-scale datasets, are further retrained on task-specific data[19-20]. This process updates the model parameters, adapting the model for improved performance on text classification tasks in particular contexts.

Another widely adopted method is feature extraction using neural network architectures. Machine learning and deep learning techniques excel in feature extraction[14,18], as different network architectures capture various linguistic features from the text and generate classification results. However, without the use of pre-training, the performance of neural network models can degrade significantly in many NLP tasks.

In addition, the local deployment and training of large language models[25,27] is becoming increasingly prevalent. These models, due to their size and complexity, are better equipped to capture the structure and semantics of language, leading to superior performance on certain tasks. However, their deployment often demands significant computational resources, posing challenges for many practical applications.

Another method frequently explored is multimodal fusion[29,33], which combines heterogeneous data sources (e.g., images, audio, and video) with text data to enhance classification accuracy. Despite its potential, this approach comes with high data collection and annotation costs, as well as the complexities associated with multimodal integration. As a result, it is often impractical for many real-world applications due to its substantial resource demands.

Therefore, in the context of text classification, it is essential to balance practical requirements, the nature of the data, and the trade-offs of different approaches to ensure efficiency and effectiveness for specific tasks.

### 2.2 Police Incident Classification

The task of police incident classification discussed in this paper represents a specialized form of text classification. Due to the urgent nature of emergency calls, coupled with the variation in communication between callers and police officers, these texts often exhibit large-scale data, complex structures, inconsistent language, and diverse levels of detail. These factors render automated classification exceptionally difficult[2]. Furthermore, because police incident data involves sensitive information related to citizens' privacy and case details, it is typically classified and unavailable to the public. As a result, most research in this domain is carried out by public security agencies and their collaborators, with limited academic investigation both domestically and internationally.

Existing research primarily employs supervised natural language processing models that integrate text preprocessing with machine learning and deep learning techniques. These methods preprocess textual data and train it within neural network frameworks to output classification labels, yielding promising results in practical applications. For instance, in 2020, Wang Mengxuan et al. enhanced the CRNN model for classifying 110 police incident texts, significantly improving classification accuracy[40]. In 2022, Qiu M and colleagues developed an intelligent model based on convolutional neural networks for police incident address recognition and classification, which improved the efficiency of incident dispatch by focusing on address information[36]. More recently, in 2024, Zhang Jing et al. introduced a BERT-DPCNN model for police incident text classification, achieving high accuracy in both binary and eleven-class classification tasks[37].

While these studies demonstrate some progress in police incident text classification, they often lack specificity in addressing particular types of incidents, especially telecom and online fraud cases, which now constitute the largest proportion of police incidents. This shortcoming hampers the ability of public security agencies to accurately extract telecom fraud cases from a wide range of incident types. To address this gap, this paper proposes and implements a KAN network algorithm integrated with a LERT-CNN-BiLSTM multichannel neural network for police incident classification. Using multi-class incident data from City B, the model achieved 91.9% accuracy in a three-class classification task (telecom fraud, non-telecom fraud, and other incidents), demonstrating its practical utility for police incident classification.

## 3 Model Construction

The overall architecture of the model constructed in this paper is shown in Fig. 1. It is roughly divided into the following parts: First, text preprocessing is based on the LERT model, which effectively extracts three types of text features and uses a masked language model to enhance contextual representation and downstream task performance. Second, a multi-channel neural network structure consisting of CNN and BiLSTM is employed, which not only captures local features of the text but also extracts contextual sequence features. Third, the KAN network is used to replace the fully connected layer structure, optimizing the neural network architecture through spline function fitting.

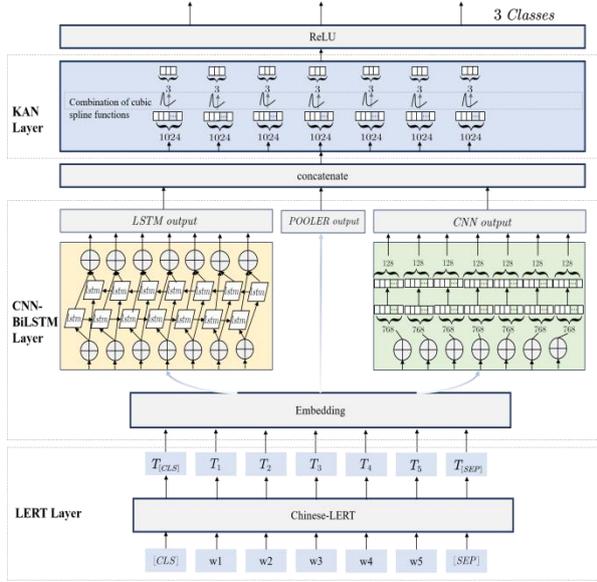

Fig. 1 Overall Model Architecture Diagram

### 3.1 LERT Preprocessing Model

LERT (Linguistically-motivated bidirectional Encoder Representation from Transformer) is a text preprocessing model that integrates multilingual features, developed in 2022 by the joint efforts of Harbin Institute of Technology and iFlytek Lab[13]. By combining WWM (Whole Word Masking) and NM (N-gram Masking) techniques, the model enhances the learning capacity of masked language models and improves the performance on downstream tasks. During feature extraction and pre-training data construction, the LERT model uses the LIP (Linguistically-Informed Pretraining) strategy to annotate language labels of the input text, extracting three types of text features: POS (Part-of-Speech), NER (Named Entity Recognition), and DEP (Dependency Parsing). Meanwhile, it employs the LTP (Language Technology Platform) to build weakly supervised pre-training data. The extracted linguistic features are then used for multi-task pre-training, alongside the original MLM (Masked Language Model) task.

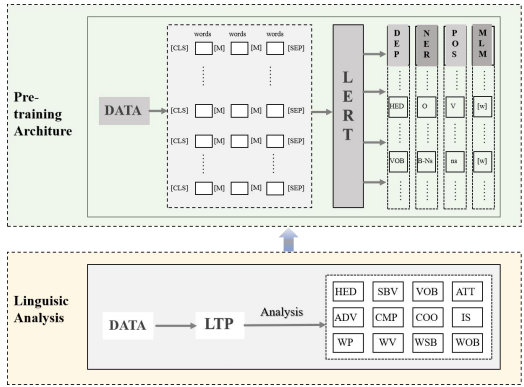

Fig. 2 LERT Model Architecture

As shown in Fig. 2, the structure of the LERT model is illustrated. The input police incident text data is first processed through LTP (Language Technology Platform), a comprehensive natural language processing platform, to analyze syntactic relations and linguistic features. The model then performs pre-training tasks, following typical BERT-like models by first tokenizing the text, adding classification (CLS) and sentence separation (SEP) markers, and applying labeling. When generating masks, LERT combines WWM and NM techniques to better capture the context and understand the language. Unlike the BERT model, which only performs the MLM task during training, the LERT model also extracts three linguistic features: DEP, NER, and POS. These tasks enhance the model's ability to understand the linguistic structure of the text. The multi-task training mode uses the LIP mechanism to allocate training weights for each task, determining the learning rate for each task in line with human cognitive understanding. After completing the pre-training tasks, the LERT model outputs an embedding layer through a fully connected layer, which is then used as input for subsequent classification models.

### 3.2 CNN and BiLSTM Network Structure

In text classification, neural network models typically perform automatic feature extraction on word embeddings generated from pre-training tasks and then compute class probabilities through a cross-entropy loss function to obtain the final classification results. CNNs (Convolutional Neural Networks) [14] and RNNs (Recurrent Neural Networks) [15] are two typical models that show excellent results in text feature extraction for classification tasks.

3.2.1 CNN Network Structure

CNN (Convolutional Neural Networks)[41] were initially applied in the field of computer vision to effectively capture both local dependencies and global information from images[42,44]. In text classification tasks, CNN traverses sentences and words using sliding windows and extracts various text features through multiple convolution filters. After the convolution layer, pooling and dimensionality reduction are typically performed, followed by fully connected output[43]. CNNs are capable of effectively capturing local features in the text, thereby improving the accuracy of text classification[45]. In this paper's task, a one-dimensional CNN is applied to the 768-dimensional text data preprocessed by LERT for convolution operations (as shown in Equation (1)). Here, X[j] represents the element at position in the input sequence, and $W_j$ represents the weight of the j-th convolution filter at position i. Each element in the input sequence is multiplied by the corresponding weight of each convolution filter at the same position, and this operation is performed across all 768 dimensions of the input sequence, followed by a weighted sum. In the end, we obtain a 128-dimensional output sequence.

$$Y_k[i] = \sum_{j=0}^{767} X[j] \times W_{k,j} \qquad (1)$$

3.2.2 BiLSTM Network Structure

BiLSTM (Bi-directional Long Short-Term Memory)[46] is a special type of RNN (Recurrent Neural Network)

structure that not only has the ability of traditional RNNs to extract sequential features but also captures bidirectional semantic dependencies and long-range relationships[47,50]. In text classification, the input text sequence is processed in both forward and backward directions, and the results are concatenated to produce the final output. This allows the model to capture information from both preceding and following time steps, thereby refining the contextual semantic information of the text[51]. The BiLSTM structure used in this paper is shown below. The 768-dimensional preprocessed sequence is input into both the forward and backward LSTM sequences, and the final output is a concatenated 128-dimensional sequence.

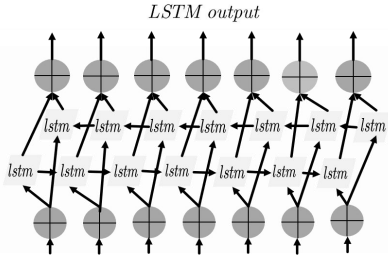

Fig. 3 BiLSTM Network Structure

To effectively leverage the strengths of both neural network structures, this paper constructs the model by parallelizing the CNN and BiLSTM network structures. This enables the extraction of both local semantic features and contextual semantic information, significantly improving the accuracy of text classification.

### 3.3 KAN Structure

The theory of KAN (Kolmogorov–Arnold Networks)[52,53] originates from the Kolmogorov-Arnold theorem, which states that any multivariable continuous function can be represented as the sum of a finite number of univariate continuous functions. Based on the core idea of this theorem, the KAN network structure no longer relies on fixed weight parameters to accommodate the model's complex relationships and structure. Instead, it utilizes more flexible spline functions for fitting[54]. Building on this, to enhance the model's representational ability and practical significance, the KAN network innovatively applies arbitrary network structures for model training, as opposed to the fixed-depth training prescribed by the original theorem. During training, the KAN network does not need to optimize the entire multivariable space. Instead, it adjusts the amount and granularity of training parameters through grids and splines, effectively reducing the number of parameters while improving the model's ability to learn fine-grained patterns.

In the natural language processing task outlined in this paper, considering the KAN network's aptitude for tasks demanding comprehension and elucidation of fundamental physical principles, we refrained from directly employing the KAN network for extracting text feature information. Rather, we substituted the linear structure of the fully connected layer with the KAN network, aiming to refine the connectivity patterns and architecture of the neural network.

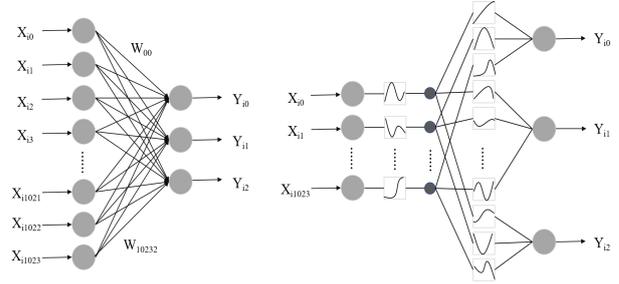

Fig. 4 Network Structure Before and After Improvement

Fig. 4 shows the network structure before (left) and after (right) improvement. After the CNN, BiLSTM, and LERT layers are concatenated, the neuron dimension becomes 1024. For the case classification task in this paper, the output needs to be a 3-dimensional classification result. When applying a fully connected layer, the output, i.e., the linear transformation of the weight matrix (which is 1024×3 in this case), is added to the bias vector. When using the KAN network, the hidden layer dimension is set to 1024, and the basic form changes as shown in the equation.

$$y_k = \sum_{j=1}^{1024} \psi_{k,j}\left(\sum_{i=1}^{1024} \phi_{j,i}(x_i)\right), \quad k=1,2,3 \qquad (2)$$

Here, $x_i$ is the i-th element of the input vector x, and $y_k$ is the k-th element of the output vector y. $\phi_{j,i}$ is the activation function from the input to the hidden layer, and $\psi_{k,j}$ is the activation function from the hidden layer to the output layer. Each activation function is represented as $\omega(b(x)+\text{spline}(x))$, where $b(x)$ is a fundamental function ($b(x)=\text{silu}(x)=x/(1+e^{-x})$ in this paper), and the spline function $\text{spline}(x)$ can be linearly calculated by multiple spline basis functions. In the text classification task in this paper, this method of reducing the number of fully connected layer parameters and minimizing accuracy loss through the KAN network structure effectively improves the performance of the deep learning model during training.

## 4 Experiment and Discussion

### 4.1 Dataset

This experiment utilizes the 2023 telecom and online fraud crime report data from City B. The data was entered by grassroots public security agencies at district-level police stations according to the standard input procedures for telecom and online fraud crime reports. It is statistically organized on a daily basis and has undergone strict de-identification processes for use in this research study. The data entry for telecom and online fraud crime reports is typically divided into two parts. One part includes information recorded by the dispatcher at the time of the emergency call, which encompasses details such as the time of the report, the victim's information (gender, age, ID number, phone number, occupation, education level, place of employment, etc.), the crime location, the type of fraud, and the amount of financial loss. The other part consists of feedback information obtained through follow-up during the investigation, which is more detailed and comprehensive. This may include the initial online platform/SMS/phone call used to contact the suspect (platform account and nickname, fraudulent phone number), communication platform details (chat account and nickname), a brief description of the fraudulent tactics, the method of transferring funds (WeChat, Alipay, online banking, etc.), additional transaction information (specific number of transfers, amounts, counterpart information, and associated bank details), and other relevant case information. A sample of the crime report data is shown in Table 1.

Table 1  Example of Police Case Text (take Fake Shopping and Services as an example)

| Summary of Police Cases |
|---|
| On December 30, 2023, at 24:00, Mr. Changju in YD Town, MTG District, B City received a strange express delivery in Building 1, Unit 101 of a certain residential area, which contained a refund notice. Mr. Changju scanned the QQ code and downloaded the Zy agent APP for the refund. He was later defrauded of 20220.8 yuan in the APP refund. The actual loss was 20220.8 yuan. On December 31, 2023, at 29:52, Mr. Changju transferred 20020 yuan to the other party (Hlj Rural Credit Cooperative, Mr. Qinglong) through (China Ping An Bank, Mr. Changju, 000000000000000000). |

According to the research needs, this study utilized a random sampling method to select 10,000 crime report cases for the experiment. The dataset was divided into training, validation, and test sets in an 8:1:1 ratio. A random seed algorithm was employed during the data split. When generating the random number sequence, the seed value 24 was used as the starting point to ensure that the same random number sequence is produced in each run, thereby ensuring the reproducibility and stability of the experimental results.

### 4.2 Experimental Environment and Parameter Settings

This study was developed using the Python programming language and the PyTorch deep learning framework. The transformers library was also utilized to load BERT and other preprocessing models. LERT and other natural language preprocessing models have been open-sourced on GitHub and can be deployed locally. Through multiple experiments, it was found that when the batch size is set to 4, the number of epochs to 3, and the learning rate to 1e-6, the model converges and reaches optimal accuracy. The experimental setup is as follows:

Table 2  Environment and settings of the experiment

| Name | Version |
|---|---|
| numpy | 1.23.5 |
| openpyxl | 3.0.10 |
| pandas | 2.1.4 |
| python | 3.11.7 |
| scikit-learn | 1.3.0 |
| tokenizers | 0.15.0 |
| torch | 2.0.0+cu118 |
| transformers | 4.36.2 |
| operating system | Win11 |
| server | NVIDIA GeForce RTX 4060 Laptop |
| batch_size | 8 |
| learning rate | 1e-6 |
| epoch | 3 |
| optimizer | Adam |

Table 3  The parameter values of the CNN-BiLSTM model

| Name | Version |
|---|---|
| convolutional dimension | 1d |
| embedding layer dimension | 768 |
| number of convolutional kernels | 128 |
| kernel_size | 1 |
| stride | 1 |
| padding | 0 |

Table 2 describes the environmental configuration for running the model, Table 3 outlines the parameter configuration for the CNN and BiLSTM models.

In the experiment, the model was evaluated using metrics such as Accuracy, Precision, Recall, F1 Score, and Average Loss on the test dataset. These metrics provide a comprehensive basis for assessing the performance of the model in classification tasks. Accuracy measures the overall correctness of the model's predictions, while Precision and Recall focus on the model's ability to distinguish between positive and negative samples. The F1 Score balances Precision and Recall, providing a more equitable evaluation of the model's performance. The

principles and formulas for calculating these metrics are as follows:

Accuracy refers to the proportion of correctly predicted samples out of the total number of samples, and its formula is:

$$Accuracy = \frac{TP+TN}{TP+TN+FP+FN} \quad (3)$$

Where TP (True Positive) is the number of true positive cases, TN (True Negative) is the number of true negative cases, FP (False Positive) is the number of false positive cases, and FN (False Negative) is the number of false negative cases.

Precision is the proportion of samples predicted as positive that are actually positive, and its formula is:

$$Precision = \frac{TP}{TP+FP} \quad (4)$$

Recall is the proportion of actual positive samples that are correctly predicted as positive by the model. It is calculated as:

$$Recall = \frac{TP}{TP+FN} \quad (5)$$

The F1 Score combines Precision and Recall into a single metric, which is their harmonic mean. It is calculated as:

$$F1\_Score = 2 \times \frac{Precision \times Recall}{Precision + Recall} \quad (6)$$

The average loss is the mean value of the loss function across all samples. It is typically used in regression tasks. In this paper, the cross-entropy loss function is used for error calculation. The principle of the cross-entropy loss function is as follows:

$$Loss = -\frac{1}{N}\sum_{i=1}^{N}(y_i \log(\hat{y}_i) + (1-y_i)\log(1-\hat{y}_i)) \quad (7)$$

Where N is the number of samples. $y_i$ is the true label of sample $i$. $\hat{y}_i$ is the predicted probability of sample being in the positive class. We calculate the average loss by averaging the loss values of all samples, which is referred to as Average Loss in this paper.

### 4.3 Experimental Results and Discussion

This study validated the components of the model through multiple sets of comparative experiments. Overall, the designed comparative experiments consist of the following parts:

1. Verifying the effectiveness of the LERT preprocessing component, which outperforms other BERT-based models in text feature extraction, enhancing the representation capability for text classification.
2. Ablation experiments to confirm the roles of the BiLSTM and CNN layers in feature extraction.
3. Comparative experiments before and after integrating the KAN network, demonstrating the advantages of the KAN network in improving the model structure.
4. Comparative experiments with other typical lightweight models, proving that the proposed model significantly outperforms existing standard text classification models.
5. Testing the robustness and generalization ability of the model by adjusting the learning rate (lr) and batch size, showcasing the rationality and scientific nature of the parameter settings.

First, the LERT preprocessing component was replaced with different BERT-based models while keeping the rest of the model structure unchanged to conduct a comparative experiment on preprocessing models. The experimental results are shown in Table 4, indicating that the LERT model performs better in the preprocessing task, with evaluation results based on five metrics outperforming other models. Specifically, accuracy, precision, recall, F1 score, and average loss are denoted as Acc, P, R, F1, and Loss, respectively.

Table 4  Comparison Experiments of the LERT Model

| Model | Acc | P | R | F1 |
|---|---|---|---|---|
| LERT-CNN-BILSTM | 0.9166667 | 0.9225357 | 0.9166667 | 0.9168044 |
| MACBERT-CNN-BILSTM | 0.9055556 | 0.9066477 | 0.9055556 | 0.9055881 |
| ROBERTA-CNN-BILSTM | 0.5466667 | 0.3986287 | 0.5466667 | 0.4469902 |
| BERT-CNN-BILSTM | 0.9144444 | 0.9195706 | 0.9144444 | 0.9135122 |

Secondly, comparative ablation experiments were designed by removing the CNN and BiLSTM layers separately to confirm their roles in feature extraction. The experimental results, as shown in Table 5, indicate that both the CNN and BiLSTM layers contribute to feature extraction, with the BiLSTM layer making a more significant contribution to improving the results. This suggests that, for the police report texts used in the experiment, it is essential to consider the entire context of the report in order to better capture the semantic information of the text.

Table 5  Comparison Experiments of the CNN-BiLSTM Model

| Model | Acc | P | R | F1 |
|---|---|---|---|---|
| LERT-CNN-BILSTM | 0.9166667 | 0.9225357 | 0.9166667 | 0.9168044 |
| LERT-CNN | 0.6588889 | 0.4977966 | 0.6588889 | 0.5502368 |
| LERT-BILSTM | 0.9100000 | 0.9141988 | 0.9100000 | 0.9099630 |

Next, to confirm the advantages of the optimized KAN network structure, we conducted comparative experiments before and after replacing the KAN network structure. The experimental results, shown in Table 6, demonstrate that the KAN network structure, through spline function fitting, improved the model's accuracy.

Table 6  Comparison Experiments Before and After Adding the KAN Network

| Model | Acc | P | R | F1 |
|---|---|---|---|---|
| LERT-CNN-BILSTM (+KAN) | 0.9188889 | 0.9255330 | 0.9188889 | 0.9188779 |
| LERT-CNN-BILSTM | 0.9166667 | 0.9225357 | 0.9166667 | 0.9168044 |

In addition, considering that other text classification models, particularly some lightweight models, might achieve superior classification performance, we conducted experiments with several commonly effective models (as shown in Fig. 5). The results indicate that these models did not achieve the desired effect.

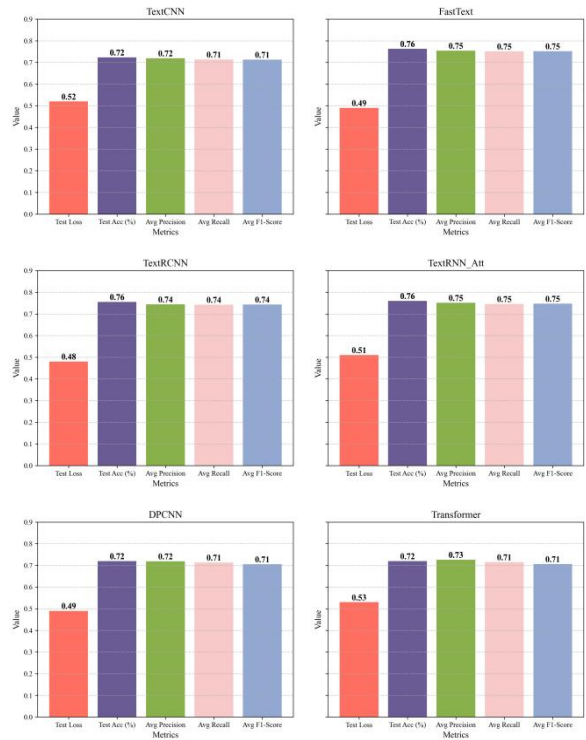

Fig. 5 Comparison Experiments Across Multiple Models

Finally, to verify the scientific nature of the experimental parameter settings, specifically whether the network could converge within a reasonable range while ensuring model performance and stability, we conducted experiments with different learning rates (lr) and batch sizes (batchsize). The results, shown in Fig. 6 and Fig. 7, demonstrate that the parameter settings in this study are reasonable, with the best performance observed when the learning rate is 1e-6 and the batch size is 4.

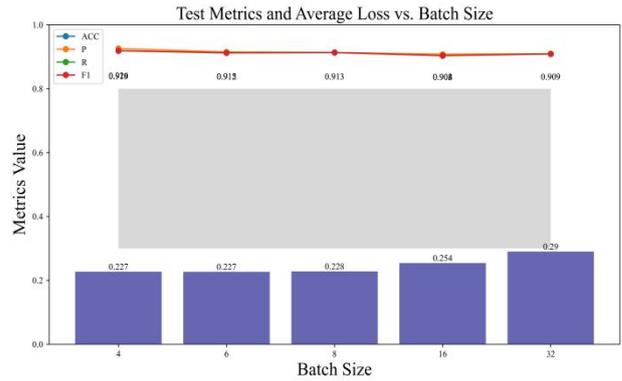

Fig. 6 Comparison Experiments with Various Batch Sizes

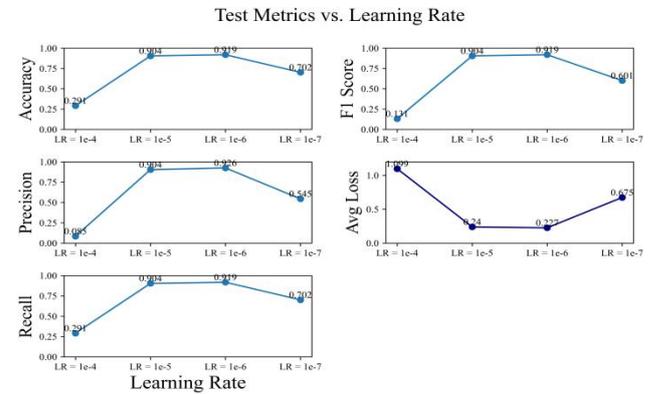

Fig. 7 Comparison Experiments with Various Learning Rates

In May 2024, based on the model architecture established in this study, we trained the model using 150,000 police report text data provided by the operational department and developed a visualization application for the automatic classification of police report texts. The classification accuracy reached 95%, significantly meeting the needs of practical applications and saving a substantial amount of police resources.

## 5  Conclusions

Intelligence information plays a crucial role in law enforcement agencies' efforts to combat and prevent illegal activities. Among these, 110 police report data represents a vast amount of information that is rich yet challenging to process. Automating the classification of this data can help law enforcement agencies implement different governance measures based on various case types, uncover potential patterns in cases, and free up human resources, thereby enhancing police efficiency. The police report classification model developed in this study, which integrates the KAN network with LERT-CNN-BiLSTM, achieves a high accuracy rate that meets the classification needs in practical law enforcement, especially for the increasingly prevalent and more complex cases of telecom and online fraud, providing an innovative classification solution. There remains potential for further transfer and application in classification scenarios in other fields. Additionally, the model requires substantial computational support, which can be optimized by pruning its structure. Most feature extraction relies on neural networks and preprocessing models, and future work can focus on more targeted text preprocessing processes, such as incorporating prompts and text augmentation.

## Acknowledgements

We would like to express our sincere gratitude to all those who have contributed to this research. First and foremost, we would like to thank the B City Public Security Bureau for providing the telecom fraud police data, which supported the completion of the experiments in this paper. Secondly, we appreciate the experimental environment provided by the People's Public Security University of China, which ensured the steady progress of the research. Additionally, we are also grateful to the Beijing Natural Science Foundation for their financial support. Finally, we extend our gratitude to all those who have contributed to this research, including my parents, teachers, classmates, boyfriend Shi Tianyang and laboratory colleagues.

## Declarations of Conflict of interest

The authors declared that they have no conflicts of interest to this work.

## Credit Statements

**Liu Zhuoxian**: Conceptualization, Methodology, Software, Writing – Original Draft **Shi Tuo**: Data Curation, Project administration, Funding acquisition **Hu Xiaofeng**: Supervision, Validation